\documentclass{article}

\usepackage{microtype}
\usepackage{graphicx}
\usepackage{subcaption}
\usepackage{booktabs} 

\usepackage{hyperref}
\usepackage{enumitem}

\usepackage[preprint]{icml2026}

\usepackage{amsmath}
\usepackage{amssymb}
\usepackage{mathtools}
\usepackage{amsthm}
\usepackage{multirow}
\usepackage{multicol}

\usepackage[capitalize,noabbrev]{cleveref}

\theoremstyle{plain}
\newtheorem{theorem}{Theorem}[section]

\theoremstyle{definition}

\theoremstyle{remark}

\usepackage[textsize=tiny]{todonotes}

\icmltitlerunning{Rethinking Multi-Modal Learning from Gradient Uncertainty}

\begin{document}

\twocolumn[
  \icmltitle{Rethinking Multi-Modal Learning from Gradient Uncertainty}



  \icmlsetsymbol{equal}{*}

  \begin{icmlauthorlist}
    \icmlauthor{Peizheng Guo}{iscas,ucas,equal}
    \icmlauthor{Jingyao Wang}{iscas,ucas,equal}
    \icmlauthor{Wenwen Qiang}{iscas,ucas}
    \icmlauthor{Jiahuan Zhou}{pku}
    \icmlauthor{Changwen Zheng}{iscas,ucas}
    \icmlauthor{Gang Hua}{amazon}
  \end{icmlauthorlist}
  
  \icmlaffiliation{iscas}{Institute of Software Chinese Academy of Sciences, Beijing, China}
  \icmlaffiliation{ucas}{University of the Chinese Academy of Sciences, Beijing, China}
  \icmlaffiliation{pku}{Wangxuan Institute of Computer Technology, Peking University, Beijing, China}
  \icmlaffiliation{amazon}{Amazon.com, Inc., Bellevue, WA, 98004, USA}

  \icmlcorrespondingauthor{Wenwen Qiang}{qiangwenwen@iscas.ac.cn}

  \vskip 0.3in
]



\printAffiliationsAndNotice{\icmlEqualContribution}  

\begin{abstract}
Multi-Modal Learning (MML) integrates information from diverse modalities to improve predictive accuracy. 
While existing optimization strategies have made significant strides by mitigating gradient direction conflicts, we revisit MML from a gradient-based perspective to explore further improvements.
Empirically, we observe an interesting phenomenon: performance fluctuations can persist in both conflict and non-conflict settings.
Based on this, we argue that: beyond gradient direction, the intrinsic reliability of gradients acts as a decisive factor in optimization, necessitating the explicit modeling of gradient uncertainty.
Guided by this insight, we propose Bayesian-Oriented Gradient Calibration for MML (BOGC-MML). 
Our approach explicitly models gradients as probability distributions to capture uncertainty, interpreting their precision as evidence within the framework of subjective logic and evidence theory. 
By subsequently aggregating these signals using a reduced Dempster's combination rule, BOGC-MML adaptively weights gradients based on their reliability to generate a calibrated update.
Extensive experiments demonstrate the effectiveness and advantages of the proposed method.

\end{abstract}

\section{Introduction}
\label{sec:introduction}
\begin{figure*}[t]
    \centering
    \includegraphics[width=\textwidth]{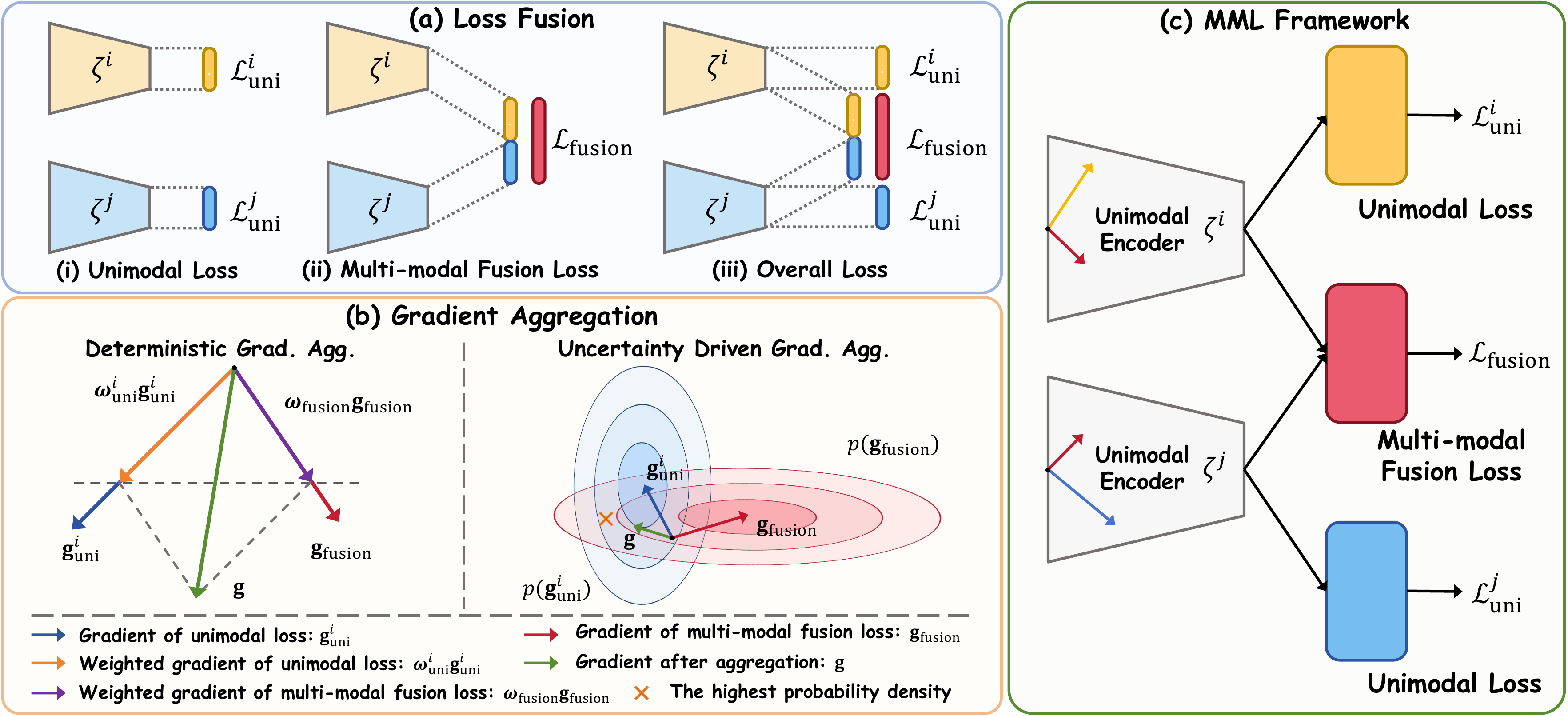}
    \caption{Illustration of the MML training framework. 
    (a) existing loss-level fusion paradigm, where unimodal losses and a multi-modal fusion loss are linearly combined. (b) gradient aggregation with fixed weights of existing methods vs. our uncertainty-driven fusion that adaptively reweights gradients based on their per-dimension reliability. (c) the overall MML pipeline. 
    }
    \label{fig:illustration}
\end{figure*}

Humans perceive the world through multiple sensory modalities, e.g., vision, hearing, and touch. This biological mechanism inspires the development of Multi-Modal Learning (MML), which aims to learn a model by integrating information from different modalities to achieve great performance \cite{huang2021makesmultimodallearningbetter,wang2023amsa,yuan2024fine,song2025bridge,ma2025multimodal}. 
Many existing approaches construct multiple unimodal losses together with a multi-modal fusion loss, and then linearly combine them for backpropagation \cite{yang2025towards,huang2022modality} (\textbf{Figure \ref{fig:illustration}}).
These frameworks have been successfully applied across various fields, e.g., healthcare \cite{muhammad2021comprehensive,shi2025heterogeneous}, autonomous systems \cite{agbley2021multimodal,feng2020deep}, human-computer interaction \cite{d2015review,li2020behrt}, and sentiment analysis \cite{poria2017review,poria2015deep,zhang2024pure}.

\begin{figure}[t]
    \centering
    \includegraphics[width=0.9\columnwidth]{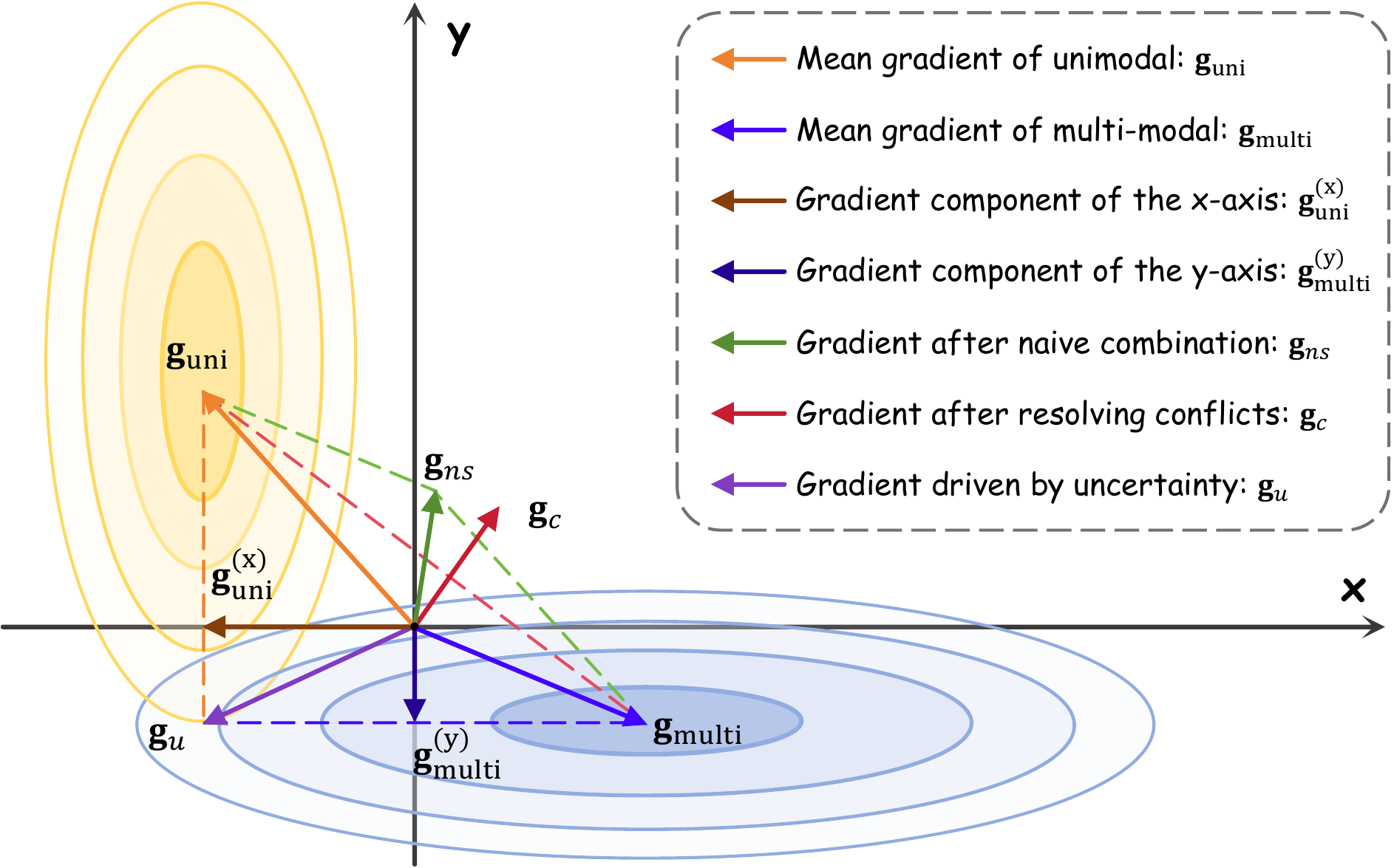}
    \caption{Illustration of different gradient aggregation.}
    \vspace{-0.2in}
    \label{fig:motivation_compare}
\end{figure}

\begin{figure}[t]
    \centering 
    
  \begin{subfigure}[b]{0.49\columnwidth}
    \centering
    \includegraphics[width=\textwidth]{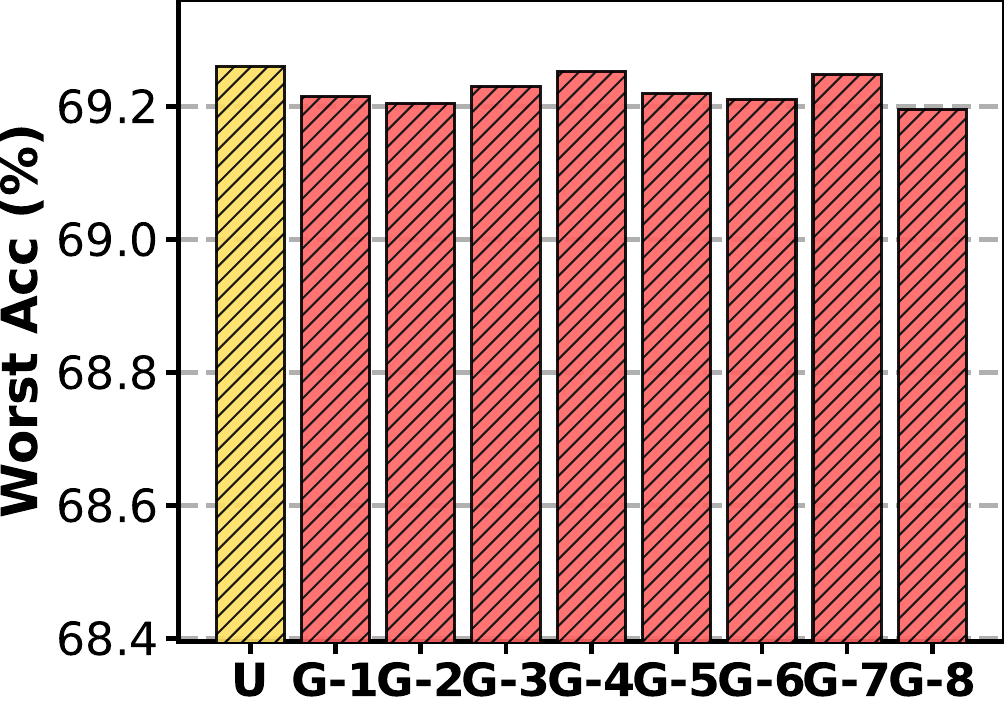}
    \caption{Conflict Scenario.}
    \vspace{-0.05in}
    \label{fig:discover_conflict}
  \end{subfigure}    
  \hfill
  \begin{subfigure}[b]{0.49\columnwidth}
    \centering
    \includegraphics[width=\textwidth]{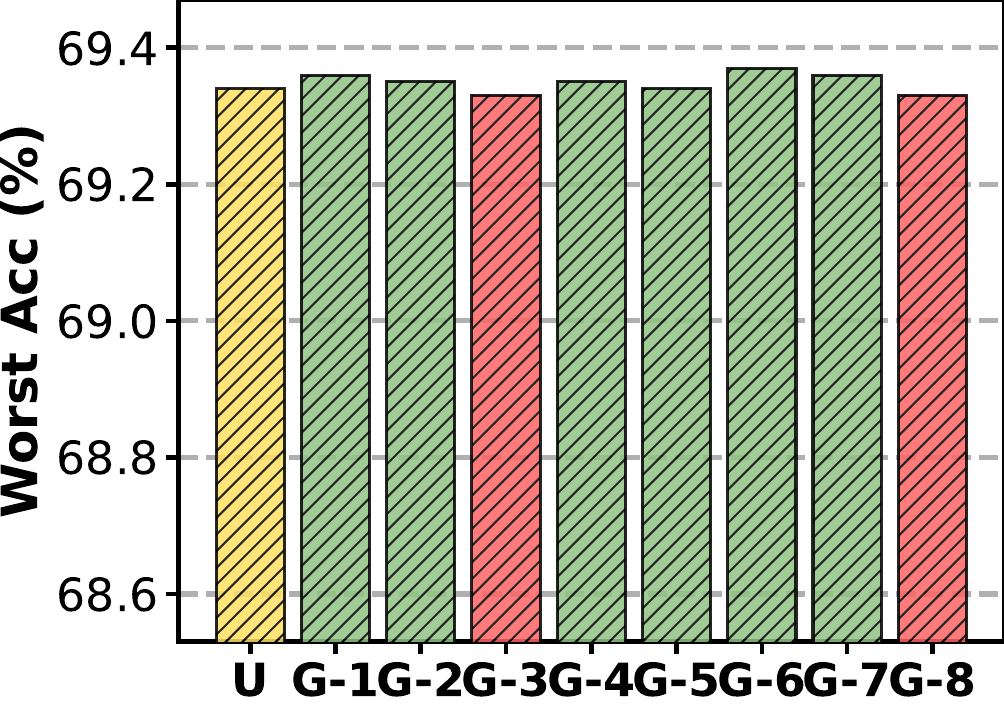}
    \caption{Non-conflict Scenario.}
    \vspace{-0.05in}
    \label{fig:discover_nonconflict}
  \end{subfigure}
    \caption{Results of the discovery experiment.}
    \vspace{-0.25in}
    \label{fig:discovery}
\end{figure}

From an optimization perspective of MML, each update is jointly driven by two gradient sources: (i) unimodal gradients derived from modality-specific objectives; and (ii) multi-modal gradients induced by fusion, alignment, or aggregation objectives (\textbf{Figure \ref{fig:illustration}b}). To enable efficient optimization, many approaches focus on the gradient directions in MML \cite{yu2020gradient,huang2022modality,chen2018gradnorm}. 
Prior works show that gradients from different objectives can be in conflict \cite{peng2022balanced,zhang2024multimodal}, and has proposed a series of methods to mitigate such conflicts \cite{wei2024enhancing,weimmpareto,wang2023large,yue2025evidential}. 
However, through empirical analysis, we observe an interesting phenomenon (\textbf{Figure \ref{fig:discovery}}): in both conflict and non-conflict settings, while mitigating gradient conflicts generally boosts performance, the model can still underperform in certain cases.

Based on the above phenomenon, we argue that gradient directions are crucial for performance, yet the reliability of gradients themselves also plays an important role in MML. We propose to analyze this from the perspective of gradient uncertainty.
A gradient is a high-dimensional vector that evolves during training, with each dimension being the partial derivative w.r.t. a model parameter \cite{goodfellow2016deep}. Unlike prior methods, we treat gradients as random variables and characterize them with probability distributions. 
Gradient uncertainty can have an important influence in gradient aggregation.
Take the two-dimensional update as an example (\textbf{Figure \ref{fig:motivation_compare}}), the unimodal gradient (yellow) has lower uncertainty along the $x$-axis but higher uncertainty along the $y$-axis, suggesting that it provides a more certain signal for updating toward the negative $x$ direction while being more uncertain for updating toward the positive $y$ direction. Similarly, the multi-modal gradient (blue) has lower uncertainty along the $y$-axis but higher uncertainty along the $x$-axis, suggesting that it is more certain for updating toward the negative $y$ direction while being more uncertain for updating toward the positive $x$ direction. In this case, both naive additive fusion and fusion after resolving conflicts can yield an update that deviates from the more certain components of the two gradients. This indicate that the more certain gradient components may be disturbed by the more uncertain ones, leading to suboptimal or even harmful update directions and thus limiting model performance. 
It also explains why performance degradation still occur under a non-conflict regime. Moreover, our experiments in \textbf{Section \ref{sec:experiment}} further show that explicitly accounting for gradient uncertainty during optimization can yield additional performance gains, highlighting that gradient reliability also plays a critical role in MML optimization. Thus, both gradient direction and gradient reliability are crucial for MML performance.
Motivated by this, we aim to incorporate gradient uncertainty into the MML optimization to potentially mitigate the performance degradation.

To this end, we propose a Bayesian-Oriented Gradient Calibration method for MML (BOGC-MML), which explicitly models the uncertainty of each gradient to achieve more effective optimization. Instead of treating the gradients as deterministic values, we model them as random variables represented by probability distributions. Specifically, inspired by Bayesian inference \cite{box2011bayesian,dempster1968generalization}, we first model the gradients as probability distributions (\textbf{Figure \ref{fig:illustration}}). Next, drawing from the ``belief-uncertainty'' in subjective logic \cite{jsang2018subjective} and evidence theory \cite{bao2021evidential,sensoy2018evidential}, we propose an efficient mapping function (\textbf{Theorem \ref{thm:precision_evidence}}) that transforms the precision of the gradient distributions into a set of evidence. Furthermore, based on the relationship between evidence, belief, and uncertainty in subjective logic, we derive the following during the gradient aggregation process: (i) belief values, representing the confidence level of the gradients at each dimension; (ii) overall uncertainty of the gradient, reflecting the reliability of this gradient as an update signal during optimization. Finally, we use a reduced Dempster's combination rule \cite{han2022trusted} to aggregate gradients, yielding a calibrated update direction that accounts for both belief and uncertainty. 
It adaptively weights gradients to amplify low-uncertainty dimensions and down-weight high-uncertainty ones, thus achieve more effective optimization.
Extensive experiments demonstrate its effectiveness.

The main contributions can be summarized as follows:
(i) We revisit MML from an optimization perspective and argue that the reliability of gradients is also a critical factor for MML through empirical analysis that model can still underperform even mitigating gradient conflicts.
(ii) We propose BOGC-MML, a novel Bayesian-oriented gradient calibration method for MML. It explicitly models the uncertainty of gradients and incorporates the dimensional differences between gradients during optimization, making the model update towards the optimal direction.
(iii) We present an effective method for estimating evidence from probability distributions with theoretical guarantees. It maps the precision of a probability distribution to its corresponding evidence, enabling more effective gradient aggregation.
(iv) Extensive experiments across various benchmarks demonstrate the effectiveness and validity of our proposed method, showcasing its advantages in practice.

\section{Related Work}
\label{sec:related_work}

Multi-modal learning (MML) aims to integrate complementary signals from modalities such as images, text, and audio to improve performance \cite{baltruvsaitis2018multimodal,ngiam2011multimodal}. A line of work tackles fusion from an optimization viewpoint by aggregating modality gradients and adaptively adjusting each modality's contribution during training \cite{peng2022balanced,tsai2019multimodal,wang2024image,wang2020makes,peng2022balanced,fan2023pmr,weimmpareto}.
For example, \cite{weimmpareto} resolves conflicts between unimodal and multi-modal objectives via Pareto-based gradient integration; \cite{yue2025evidential} uses an evidential dissonance measure to assess view quality and performs dissonance-aware belief integration with an evidential gradient penalty for robust multi-view fusion; \cite{jiang2025interactive} modifies gradients via a flat gradient modification strategy (with SAM-style smoothing) to enhance cross-modal interaction and alleviate modality imbalance.
However, these methods often rely on fixed coefficients and implicitly treat gradients as equally reliable, overlooking gradient uncertainty and dimension-wise sensitivity. We address this by explicitly modeling each modality gradient as a distribution and steering fusion toward more reliable update directions. More discussions are provided in \textbf{Appendix C}.

\section{Problem Settings and Analyses} 
In this section, we first formalize important concepts and theories in this work. Subsequently, we investigate the impact of gradient reliability on model performance through empirical analysis, which also motivates our proposed method.

\subsection{Problem Settings}
\label{sec:preliminary}
In this subsection, we introduce the problem settings and related theories of this work. For more details and illustration of relevant concepts and theories, e.g., subjective logic and evidence theory, dempster-shafer evidence theory, etc., are provided in \textbf{Appendix C.1}.

\textbf{Multi-Modal Learning} 
In MML, the goal is to make accurate predictions by effectively integrating information from multiple modalities. To achieve this, it is common to jointly optimize via multi-modal fusion loss and unimodal losses  \cite{wang2020makes} (\textbf{Figure \ref{fig:illustration}a}). The former is computed based on the combined representation of all modalities and directly guides the model's predictions; the latter supervises the learning of each modality representation, encouraging the model to extract important features from each modality.

Given a dataset $D^i = \{(x_t^{i}, y_t)\}_{t=1}^n$ where $x_t^{i}$ represents the input of the $i$-th modality for the $t$-th sample, and $y_t$ is the corresponding label. The overall loss function is defined as:
\begin{equation}\label{eq3.1.1}
\mathcal{L}_{\rm total} = \mathcal{L}_{\rm fusion} + \varphi {\textstyle\sum_{i=1}^{M}} \mathcal{L}_{\rm uni}^{i},
\end{equation}
where $\mathcal{L}_{\rm fusion}$ is the multi-modal fusion loss,  $\mathcal{L}_{\rm uni}^{i}$ is the unimodal loss for the $i$-th modality, $\varphi$ is a regularization parameter, and $M$ is the total number of modalities.

\subsection{Empirical Evidence and Motivation}
Existing MML methods typically sum unimodal gradients and multi-modal gradients or mix them with fixed ratios during optimization \cite{yu2020gradient,wang2020makes,huang2022modality,weimmpareto,wei2024enhancing}. However, each update of unimodal encoders is jointly driven by two gradient sources: 
(i) unimodal gradients from unimodal losses, and (ii) multi-modal gradients from fusion. Both sources usually fluctuate across mini-batches and training stages, and a fixed fusion rule may not guarantee consistently beneficial updates, affecting performance. We further conduct an experiment to investigate this phenomenon.

Specifically, we adopt the Uniform baseline as the reference without any additional gradient aggregation strategy, and use G-Blending \cite{wang2020makes} as a representative method that fuses unimodal gradients with multi-modal fusion gradients during updates. We partition training scenarios into non-conflict and conflict cases and evaluate the final performance separately for each case. We further repeat G-Blending with the same configuration and report Worst Acc on CREMA-D \cite{cao2014crema}.
As shown in \textbf{Figure \ref{fig:discover_nonconflict}}, under the non-conflict setting, G-Blending slightly outperforms Uniform in most repeats (green bars above the yellow bar), but still underperforms in a few runs (red bars). In the conflict setting (\textbf{Figure \ref{fig:discover_conflict}}), overall performance is lower and most G-Blending runs show larger drops than Uniform, although a small fraction still achieve marginal gains. Overall, these results suggest that while mitigating gradient conflicts can generally improve performance, the model can still underperform in certain cases.

Observing these results, we claim that gradient directions are essential for performance, but the reliability of gradients also has a significant impact on MML. A gradient can be viewed as a high-dimensional vector evolving throughout training, and each dimension is the partial derivative w.r.t. a model parameter \cite{goodfellow2016deep}. We model gradients as random variables and describe them by probability distributions. We take two-dimensional updates as an example for further explanation (\textbf{Figure \ref{fig:motivation_compare}}). Uncertainty can be heterogeneous across dimensions and sources: the unimodal gradient (yellow) is more certain in the $x$-axis to update toward the negative $x$ direction, and the multi-modal gradient (blue) in the $y$-axis to update toward the negative $y$ direction. However, either by naive additive fusion or fusion after resolving conflicts, the obtained update differs from the more certain components of the two gradients, suggesting that uncertain gradient components may distort components where the other is reliable. Consequently, the update may be suboptimal or even harmful, leading to limited performance. This corresponds to the phenomenon we observed. In \textbf{Section \ref{sec:experiment}}, the experiments suggest that taking into account uncertainty during optimization can bring further improvement in model performance. 
Thus, not only the gradient directions are important, but also the reliability of the gradients play a significant role for MML. 
More details and results are provided in \textbf{Appendix G}.
Motivated by this insight, we propose an uncertainty-aware method for gradient aggregation for efficient optimization.

\section{Methodology}
\label{sec:methodology}
Based on the above analyses, we propose BOGC-MML, a Bayesian-oriented gradient calibration method for MML. It calibrates the update direction during MML optimization by incorporating the uncertainty of each gradient.
Specifically, we first model the gradient of each modality as a probability distribution using Bayesian inference. We then derive the gradient distribution for the multi-modal fusion loss (\textbf{Section \ref{sec:gradient_distribution}}).
Next, we propose an effective method to estimate evidence from these probability distributions, i.e., using a power mapping that is based on the precision of the probability distributions (\textbf{Theorem \ref{thm:precision_evidence}}).
Obtaining the estimated evidence, we quantify the uncertainties of the modality-specific gradients and the gradient of the multi-modal fusion loss, drawing from subjective logic and evidence theory.
Finally, based on the quantified uncertainties, we perform uncertainty-driven gradient aggregation using the reduced Dempster's combination rule to obtain a more precise update direction, calibrating MML optimization  (\textbf{Section \ref{sec:gradient_aggregation}}).

\subsection{Bayesian-based Gradient Probability Distribution}
\label{sec:gradient_distribution}

In this subsection, we quantify modality-gradient uncertainty by inferring their probability distributions using Bayesian inference. 
We treat the parameters of the last pre-output layer as random variables and estimate their posteriors using a Laplace approximation \cite{mackay1992bayesian}. 
Then, we express each modality gradient as a quadratic function of these parameters and approximate its distribution via first- and second-order moments.
Finally, given per-modality gradient distributions, we derive the gradient distribution of the multi-modal fusion loss under Gaussian priors.

\textbf{Posterior Estimation of Parameters.}
The network architectures of MML typically consist of modality-specific encoders, modality fusion layers, and classification heads \cite{baltruvsaitis2018multimodal,ngiam2011multimodal}.
We consider the parameters $\boldsymbol{\Theta}$ of the last layer before output as random variables and compute their probability distributions. Assuming that the posterior distribution of $\boldsymbol{\Theta}$ is a Gaussian distribution, which is a mild assumption \cite{neal2012bayesian}, we approximate the posterior using the Laplace approximation \cite{mackay1992bayesian}. Given the observed data $\mathcal D$, the posterior distribution can be formulated through Bayesian rule $p(\boldsymbol{\Theta}|\mathcal D)\propto p(\mathcal D|\boldsymbol{\Theta}) \cdot  p(\boldsymbol{\Theta})$.
Then, we approximate the posterior distribution of the parameters via a second-order Taylor expansion and Laplace approximation: for the $i$-th modality, the posterior distribution of $\boldsymbol{\Theta}^i$ is: 
\begin{equation}\label{eq4.1.2}
\resizebox{\linewidth}{!}{$
\begin{aligned}
\log p(\boldsymbol{\Theta}^i\mid\mathcal{D}) &\approx  \log p(\hat{\boldsymbol{\Theta}}^i\mid\mathcal{D})+(-\nabla_{\boldsymbol{\Theta}^i} \log p(\mathcal{D}, \boldsymbol{\Theta}^i))^\top \Delta \boldsymbol{\Theta}^i 
\\&+ \frac{1}{2} (\Delta \boldsymbol{\Theta}^i)^\top \mathbf{H}^i \Delta \boldsymbol{\Theta}^i,
\end{aligned}
$}
\end{equation}
where $\hat{\boldsymbol{\Theta}}^i$ is the learned point estimate of the parameters, $\Delta \boldsymbol{\Theta}^i = \boldsymbol{\Theta}^i-\hat{\boldsymbol{\Theta}}^i$ is the deviation from the estimate, $\mathbf{q}=-\nabla_{\boldsymbol{\Theta}} \log p(\mathcal{D}, \boldsymbol{\Theta})$, and $\mathbf{H}^i=-\nabla_{\boldsymbol{\Theta}^i}^2 \log p(\mathcal{D}, \boldsymbol{\Theta}^i)$ is the negative Hessian matrix of the joint log-likelihood with respect to $p(\mathcal{D}, \boldsymbol{\Theta}^i)$.
Then, we omit the first-order term (i.e. $(-\nabla_{\boldsymbol{\Theta}^i} \log p(\mathcal{D}, \boldsymbol{\Theta}^i))^\top \Delta \boldsymbol{\Theta}^i$) in the Taylor expansion. As a result, the posterior takes the quadratic form: $\log p(\boldsymbol{\Theta}^i\mid\mathcal{D}) \approx c^i + \frac{1}{2}(\Delta \boldsymbol{\Theta}^i)^\top \mathbf{H}^i \Delta \boldsymbol{\Theta}^i$, where $c^i$ absorbs terms independent of $\boldsymbol{\Theta}^i$. 
The corresponding Gaussian distribution is (with details in \textbf{Appendix C.3}):
\begin{equation}\label{eq4.1.3}
p(\boldsymbol{\Theta}^i\mid\mathcal{D}) \approx \mathcal{N}(\boldsymbol{\Theta}^i \mid (\hat{\boldsymbol{\Theta}}^i-(\mathbf{H}^i)^{-1}\mathbf{q}^i), (\mathbf{H}^i)^{-1}).
\end{equation}

However, the Hessian $\mathbf{H}^i$ is often not positive-definite \cite{achituve2024bayesian,choromanska2015loss,dauphin2014identifying}. We replace it with the generalized Gauss-Newton matrix, which provides a more stable approximation:
\begin{equation}\label{eq4.1.4}
\tilde{\mathbf{H}}^i = \sum_{j=1}^{|\mathcal D_{batch}|} (\mathbf{J}_j^i)^\top \mathbf{B}_j^i \mathbf{J}_j^i + (\mathbf{V}_p^i)^{-1},
\end{equation}
where $|\mathcal D_{batch}|$ is the batch size, $\mathbf{J}_j^i = \nabla_{\boldsymbol{\Theta}^i} \mathbf{f}^i(X_j; \boldsymbol{\Theta}^i)$ is the Jacobian of the model output for the $i$-th modality with respect to its parameters, $\mathbf{B}_j^i = \nabla_{\mathbf{f}^i}^2 (-\log p(Y_j^i \mid X_j, \boldsymbol{\Theta}^i))$ is the Hessian of the negative log-likelihood with respect to the model outputs, and $\mathbf{V}_p^i$ is the covariance of the Gaussian prior for $\boldsymbol{\Theta}^i$.
Thus, the posterior approximation is:
\begin{equation}\label{eq4.1.5}
p(\boldsymbol{\Theta}^i\mid\mathcal{D}) \approx \mathcal{N}(\boldsymbol{\Theta}^i \mid (\hat{\boldsymbol{\Theta}}^i-(\mathbf{H}^i)^{-1}\mathbf{q}^i), \tilde{\mathbf{H}}^i).
\end{equation}

\textbf{Gradient Distribution of Single Modality.}
Obtaining the posterior distributions, we then approximate the gradient probability distribution of the modalities. Specifically, we first establish the functional dependence of the gradient on the parameters. Next, we compute the first and second-order moments of the gradient through Monte Carlo sampling. 
Finally, the gradient distribution is approximated by a Gaussian distribution, from which the corresponding mean and covariance are derived.

Given the $i$-th modality and a random batch of samples $\mathcal D_{batch}\sim \mathcal D$, the gradient of the square loss with respect to the hidden layer $\mathbf{\psi}$ for the $t$-th sample is derived as:
\begin{equation}\label{eq4.1.7}
\mathbf{g}_{t}^i=\frac{\partial \mathcal{L}_{t}^i}{\partial \hat{Y}_{t}^i}\frac{\partial \hat{Y}_{t}^i}{\partial \mathbf{\psi}_{t}}=2\boldsymbol{\Theta}^i\mathbf{\psi}_{t}^\top\boldsymbol{\Theta}^i-2\boldsymbol{\Theta}^iY_{t}^i,
\end{equation}
where $\mathcal{L}_{t}^i=(Y^i-\hat{Y}^i)^2$ is the square loss between label $Y^i$ and network output $\hat{Y}^i$ for modality \textit{i}.

\textbf{Eq.~\ref{eq4.1.7}} defines the functional relationship between the gradient and the model parameters. Specifically, $\mathbf{g}_{t}^i$ is a quadratic function of the random variable $\boldsymbol{\Theta}^i$, which follows a Gaussian distribution. 
Consequently, $\mathbf{g}_{t}^i$ is also a random variable, and it follows a generalized chi-square distribution \cite{davies1973numerical}. 
The true distribution of $\mathbf{g}_{t}^i$ is generally intractable, as it does not admit a closed-form expression for its probability density \cite{bishop2006pattern,jordan1999introduction}. To enable efficient computation and inference, we approximate it with a Gaussian distribution \cite{rice2007mathematical}. This choice is motivated by the fact that the Gaussian distribution has a closed-form density and is determined by its mean and covariance, making it analytically convenient \cite{bishop2006pattern,murphy2012machine}. In practice, we estimate the parameters of this Gaussian approximation by matching the first- and second-order moments of $\mathbf{g}_{t}^i$, following the moment-matching principle \cite{minka2001family}.

In classification tasks, the gradient exhibits a nonlinear dependency on the model parameters, which precludes a closed-form solution for the first- and second-order moments \cite{bishop2006pattern,williams2006gaussian}.
To address this, we first approximate the gradient distribution by assuming a Gaussian distribution, and then match its first- and second-order moments to make it as close as possible to the true distribution. Since directly obtaining these moments from the true distribution is intractable \cite{bishop2006pattern,welling2011bayesian}, we use Monte Carlo sampling \cite{robert1999monte} to approximate them. Monte Carlo sampling only requires drawing samples from the posterior distribution and computing the gradient for each sample, without needing to rely on the specific structure of the model \cite{robert1999monte, welling2011bayesian}.
Thus, given $n$ samples, we can derive the mean and covariance of the Gaussian approximation as follows:
\begin{equation}\label{eq4.1.8}
\resizebox{\linewidth}{!}{$
\begin{aligned}
&\boldsymbol{\mu }_{t}^i:=\mathbb{E}[\mathbf{g}^i_{t}]\approx \frac{1} {n}\sum_{j=1}^{n}\mathbf{g}_{t}^i(\boldsymbol{\Theta}_j^i),\quad\\
&\boldsymbol{\Sigma }_{t}^i:=\mathbb{E}[(\mathbf{g}^i_{t})^2]-\boldsymbol{\mu }_{t}^i(\boldsymbol{\mu }_{t}^i)^\top \approx \frac{1}{n}\sum_{j=1}^{n}\mathbf{g}_{t}^i(\boldsymbol{\Theta}_j^i)\mathbf{g}_{t}^i(\boldsymbol{\Theta}_j^i)^\top-\boldsymbol{\mu }_{t}^i(\boldsymbol{\mu }_{t}^i)^\top,
\end{aligned}
$}
\end{equation}
where $\boldsymbol{\Lambda}_t^i=\frac{1}{\boldsymbol{\Sigma }_{t}^i}$ is the precision matrix, $\boldsymbol{\Sigma }_{t}^i$ and $\boldsymbol{\Lambda}_t^i$ are both $K$-dimensional matrices, and $\boldsymbol{\Theta}_j^i$ are samples from $p(\boldsymbol{\Theta}^i|\mathcal D_{batch})$.
The probability distribution of $\mathbf{g}_{t}^i$ is:
\begin{equation}\label{eq4.1.9}
    p(\mathbf{g}_{t}^i)\approx \mathcal{N}(\mathbf{g}_{t}^i\mid\boldsymbol{\mu }_{t}^i,\boldsymbol{\Sigma }_{t}^i).
\end{equation}

\textbf{Gradient Distribution of the Multi-Modal Fusion Loss.}
To obtain the gradient probability distribution of $\mathcal{L}_{\rm fusion}$, we proceed in two steps: (i) derive the expression of the gradient of $\mathcal{L}_{\rm fusion}$, and (ii) calculate the gradient distribution based on the Gaussian priors \cite{papoulis1965random}.

For the first step, in MML, the fusion loss $\mathcal{L}_{\rm fusion}$ can be expressed as $\mathcal{L}_{\rm fusion}=\mathcal{L}(\mathcal{C}(\zeta^1\oplus\zeta^2\oplus\cdots\oplus\zeta^m),y)$, 
where each modality is processed by a different modality-specific encoder $\zeta^i$ with parameter $\theta^i$ (\textbf{Figure \ref{fig:illustration}}), $\mathcal{C}(\cdot)$ is a classifier, and $\oplus$ denotes a fusion operation (e.g., concatenation). 
Given the $t$-th sample, we obtain the the gradient of $\mathcal{L}_{\rm fusion}$, i.e., $\mathbf{g}_{t}^{\rm fusion}=\frac{1}{M}\sum_{i=1}^M \mathbf{g}_{t}^i$.
Furthermore, since the Gaussian distribution remains Gaussian after weighted sum \cite{papoulis1965random}, we can obtain the mean and covariance of $\mathbf{g}_{t}^{\rm fusion}$ based on the gradient probability distributions of each modality, which can be expressed as:
\begin{equation}\label{eq4.2.1}
\begin{aligned}
    \scalebox{0.95}{$\boldsymbol{\mu }_{t}^{\rm fusion}=\frac{1}{M}\sum_{i=1}^M\boldsymbol{\mu }_{t}^i,\quad
    \boldsymbol{\Sigma }_{t}^{\rm fusion}=\frac{1}{M}\sum_{i=1}^M\boldsymbol{\Sigma }_{t}^i,$}
\end{aligned}
\end{equation}
where $\boldsymbol{\Sigma }_{t}^{\rm fusion}$ is also a $K$-dimensional matrix.
Thus, the probability distribution of $\mathbf{g}_{t}^{\rm fusion}$ is:
\begin{equation}\label{eq4.2.2}
    p(\mathbf{g}_{t}^{\rm fusion})\approx \mathcal{N}(\mathbf{g}_{t}^{\rm fusion}\mid\boldsymbol{\mu }_{t}^{\rm fusion},\boldsymbol{\Sigma }_{t}^{\rm fusion}).
\end{equation}
In summary, in this subsection, we first capture the uncertainty of each modality by modeling its gradient distribution through the posterior distribution. 
This approach yields a probabilistic description of the individual gradient spaces. 
Building on these per-modality distributions, we derive the Gaussian distribution for the fused gradient $\mathbf{g}_t^{\rm fusion}$ of the multi-modal fusion loss $\mathcal{L}_{\rm fusion}$.

\begin{table*}[t]
\centering
\caption{Classification accuracy(\%) on CREMA-D and Kinetics Sounds. The best results are shown in \textbf{bold}. * indicates that the unimodal accuracy (Acc audio and Acc video) is obtained by fine-tuning a unimodal classifier with the trained unimodal encoder frozen.}
\vspace{-0.05in}
\label{tab:imbalance_cnn}
\setlength{\tabcolsep}{1.8mm}
\resizebox{\linewidth}{!}{
\begin{tabular}{l|cccc|cccc}
\bottomrule
\multirow{2}{*}{\textbf{Method}} & \multicolumn{4}{c|}{\textbf{CREMA-D}} & \multicolumn{4}{c}{\textbf{Kinetics Sounds}} \\
  & \textbf{Acc} & \textbf{Acc audio} & \textbf{Acc video} & \textbf{Worst Acc} & \textbf{Acc} & \textbf{Acc audio} & \textbf{Acc video} & \textbf{Worst Acc} \\ \hline
Audio-only & - & 61.69 & - & - & - & 53.63 & - & - \\
Video-only & - & - & 56.05 & - & - & - & 49.20 & - \\
Unimodal pre-trained \& fine-tune & 71.51 & 60.08 & \textbf{60.22} & 68.51 & 68.75 & 53.49 & 50.07 & 66.75 \\
One joint loss* & 66.13 & 59.27 & 36.56 & 63.63 & 64.61 & 52.03 & 35.47 & 61.61 \\
Uniform baseline & 71.10 & 63.44 & 51.34 & 69.30 & 68.31 & 53.20 & 40.55 & 66.11 \\ \hline
G-Blending \cite{wang2020makes} & 72.01 & 60.62 & 52.23 & 69.31 & 68.90 & 52.11 & 41.35 & 65.40 \\
OGM \cite{peng2022balanced}* & 69.19 & 56.99 & 40.05 & 66.89 & 66.79 & 51.09 & 37.86 & 64.90 \\
Greedy \cite{wu2022characterizing}* & 67.61 & 60.69 & 38.17 & 65.51 & 65.32 & 50.58 & 35.97 & 62.52 \\
PMR \cite{fan2023pmr}* & 66.32 & 59.95 & 32.53 & 64.62 & 65.70 & 52.47 & 34.52 & 62.30 \\
AGM \cite{li2023boosting}* & 70.06 & 60.38 & 37.54 & 68.06 & 66.17 & 51.31 & 34.83 & 63.67 \\
TMC \cite{han2022trusted} & 72.84 $\pm$ 0.98 & 61.73 $\pm$ 1.02 & 52.25 $\pm$ 0.88 & 69.76 $\pm$ 0.93 & 70.17 $\pm$ 1.12 & 55.53 $\pm$ 0.86 & 50.83 $\pm$ 1.21 & 67.64 $\pm$ 0.82 \\
MMPareto \cite{weimmpareto} & 75.13 & 65.46 & 55.24 & 71.93 & 70.13 & 56.40 & 53.05 & 67.33 \\ 
D\&R \cite{wei2024diagnosing} & 73.52 & 61.94 & 52.59 & 70.13 & 69.10 & 54.18 & 44.86 & 66.75 \\ 
ARL \cite{wei2025improving} & 74.19 & 62.15 & 52.91 & 70.50 & 68.40 & 53.10 & 40.70 & 66.01 \\ 
ECML \cite{yue2025evidential} & 74.68 $\pm$ 1.17 & 62.31 $\pm$0.95 & 53.16 $\pm$ 0.83 & 70.77 $\pm$ 1.07 & 71.25 $\pm$ 1.17 & 56.24 $\pm$ 0.95 & 51.35 $\pm$ 1.17 & 68.08 $\pm$ 0.87 \\
\hline
BOGC-MML & \textbf{78.91 $\pm$ 0.78} & \textbf{68.52 $\pm$ 0.85} & 58.79 $\pm$ 0.59 & \textbf{74.02 $\pm$ 0.76} & \textbf{72.64 $\pm$ 0.82} & \textbf{58.11 $\pm$ 0.79} & \textbf{55.89 $\pm$ 0.91} & \textbf{69.37 $\pm$ 0.65} \\ 
$\Delta$ (v.s. \text{SOTA}) & $\uparrow$ 3.78 & $\uparrow$ 3.06 & $\downarrow$ 1.43 & $\uparrow$ 2.09 & $\uparrow$ 1.39 & $\uparrow$ 1.71 & $\uparrow$ 2.84 & $\uparrow$ 1.29 \\ 
\toprule
\end{tabular}}
\vspace{-0.05in}
\end{table*}

\subsection{Uncertainty Driven Gradient Aggregation}
\label{sec:gradient_aggregation}

Based on the above results, we aggregate the gradient $\mathbf{g}_{t}^i$ of each modality with the gradient $\mathbf{g}_{t}^{\rm fusion}$ of the multi-modal loss, taking into account the uncertainty in each modality. The aggregated results are used to determine the optimal update direction. 
Specifically, inspired by subjective logic \cite{jsang2018subjective} and evidence theory \cite{bao2021evidential,sensoy2018evidential}, we propose an effective method to convert each gradient into Dirichlet concentration parameters and associated evidence values. Obtaining these results, we compute the belief masses and uncertainty for each gradient in every dimension. Next, we fuse these beliefs and uncertainties across the gradients using the reduced Dempster's combination rule, yielding a joint belief and a joint uncertainty per dimension. Finally, we weight and sum $\mathbf{g}_{t}^i$ and $\mathbf{g}_{t}^{\rm fusion}$ according to the fused belief and uncertainty, producing the calibrated update direction for the $i$-th modality.

Given the $i$-th modality and $t$-th sample, the covariance matrix $\boldsymbol{\Sigma} ^i_t$ and the precision matrix $\boldsymbol{\Lambda}_t^i$ are both $K$-dimension (\textbf{Eq.\ref{eq4.1.8}}). The element of the $d$-th row and $d$-th column in $\boldsymbol{\Lambda}_t^i$ is the precision of $\mathbf{g}_{t}^i$ in the $d$-th dimension. Similarly, the precision of $\mathbf{g}_{t}^{\rm fusion}$ in the $d$-th dimension can be obtained according to \textbf{Eq.\ref{eq4.2.2}}, i.e., $\boldsymbol{\lambda }_{t,d}^i=(\boldsymbol{\Sigma} ^i_t)^{-1}[d,d]$ and $\boldsymbol{\lambda }_{t,d}^{\rm fusion}=(\boldsymbol{\Sigma} ^{\rm fusion}_t)^{-1}[d,d]$. Then, we propose:
\begin{theorem}
\label{thm:precision_evidence}
Given a Gaussian distribution $\theta \sim\mathcal{N}\bigl(\mu_0,\;\tfrac{1}{\lambda_0}\bigr)$. Given $n$ observations $x_1,\dots,x_n$ and the observation noise precision (inverse variance) is $\tau>0$, the likelihood is $p(x_{1:n}|\theta)=\prod_{i=1}^n\mathcal{N}(x_i|\theta,\tfrac{1}{\tau})$. Given a real number exponent $s>0$ and a mapping $\phi\colon (0,\infty)\to(0,\infty)$ where $\phi(\lambda)=\lambda^s$, its scalar evidence can be estimated as $e =\phi(\lambda_n)\;=\bigl(\lambda_0 + n\tau\bigr)^s$.
\end{theorem}
\textbf{Theorem \ref{thm:precision_evidence}} provides a method to convert the parameter posterior precision, which is obtained by combining a Gaussian prior with observed data into a single evidence value via a continuous monotonic mapping. The prior and data merge into a posterior precision (inverse variance), and then an exponent-controlled mapping translates that precision into evidence. The resulting evidence both captures the amount of information and allows us to adjust how sensitively it responds to changes in precision, unifying prior knowledge and observed uncertainty into one coherent measure.

\begin{table*}[t]
\centering
\caption{Test performance on CREMA-D and Kinetics Sounds where \textbf{bold} and underline represent the best and runner-up respectively. The network follows transformer-based framework MBT \cite{nagrani2021attention}.}
\vspace{-0.05in}
\label{tab:imbalance_mbt}
\setlength{\tabcolsep}{2mm}
\resizebox{\linewidth}{!}{
\begin{tabular}{l|cc|cc|cc|cc}
\bottomrule
\multirow{3}{*}{\textbf{Method}} & \multicolumn{4}{c|}{\textbf{CREMA-D}} & \multicolumn{4}{c}{\textbf{Kinetics Sounds}} \\
& \multicolumn{2}{c}{\textbf{from scratch}} & \multicolumn{2}{c|}{\textbf{with pretrain}} & \multicolumn{2}{c}{\textbf{from scratch}} & \multicolumn{2}{c}{\textbf{with pretrain}} \\
\cmidrule(lr){2-3} \cmidrule(lr){4-5} \cmidrule(lr){6-7} \cmidrule(lr){8-9}
& \textbf{Acc} & \textbf{macro F1} & \textbf{Acc} & \textbf{macro F1}  & \textbf{Acc} & \textbf{macro F1} & \textbf{Acc} & \textbf{macro F1} \\ \hline
One joint loss & 44.96 & 42.78 & 66.69 & 67.26 & 42.51 & 41.56 & 68.30 & 69.31 \\
Uniform baseline & 45.30 & 43.74 & 69.89 & 70.11 & 43.31 & 43.08 & 69.40 & 69.60 \\ \hline
G-Blending \cite{wang2020makes} & 46.38 & 45.16 & 69.91 & 70.01 & 44.69 & 44.19 & 69.41 & 69.47 \\
OGM-GE \cite{peng2022balanced} & 42.88 & 39.34 & 65.73 & 65.88 & 41.79 & 41.09 & 69.55 & 69.53 \\
Greedy \cite{wu2022characterizing} & 44.49 & 42.76 & 66.67 & 67.26 & 43.31 & 43.08 & 69.62 & 69.75 \\
PMR \cite{fan2023pmr} & 44.76 & 42.95 & 65.59 & 66.07 & 43.75 & 43.21 & 69.67 & 69.87 \\ 
AGM \cite{li2023boosting} & 45.36 & 43.81 & 66.54 & 67.75 & 43.65 & 43.57 & 69.59 & 69.14 \\ 
TMC \cite{han2022trusted} & 46.83 $\pm$ 0.83 & 46.26 $\pm$ 0.93 & 68.96 $\pm$ 0.85 & 69.61 $\pm$ 1.02 & 43.91 $\pm$ 0.79 & 43.95 $\pm$ 0.84 & 69.02 $\pm$ 0.69 & 68.56 $\pm$ 0.72 \\
ECML \cite{yue2025evidential} & 48.00 $\pm$ 0.78 & 47.26 $\pm$ 0.97 & 70.11 $\pm$ 0.69 & 70.86 $\pm$ 0.95 & 45.13 $\pm$ 0.74 & 44.75 $\pm$ 0.93 & 70.12 $\pm$ 0.82 & 69.61 $\pm$ 0.92 \\
MMPareto \cite{weimmpareto} & 48.66 & 48.17 & 70.43 & 71.17 & 45.20 & 45.26 & 70.28 & 70.11 \\ \hline
BOGC-MML & \textbf{52.21 $\pm$ 0.69} & \textbf{50.06 $\pm$ 0.74} & \textbf{74.00 $\pm$ 0.78} & \textbf{73.01 $\pm$ 0.83} & \textbf{47.19 $\pm$ 0.68} & \textbf{47.36 $\pm$ 0.86} & \textbf{73.17 $\pm$ 0.79} & \textbf{73.08 $\pm$ 0.83} \\ 
$\Delta$ (v.s. \text{SOTA}) & $\uparrow$ 3.55 & $\uparrow$ 1.89 & $\uparrow$ 3.57 & $\uparrow$ 1.84 & $\uparrow$ 1.99 & $\uparrow$ 2.10 & $\uparrow$ 2.89 & $\uparrow$ 2.97 \\
\toprule
\end{tabular}}
\end{table*}

\begin{table*}[t]
\caption{Performance comparison when 50\% samples are corrupted with Gaussian noise, i.e., zero mean with the variance of $N$. ``(N, Avg.)'' and ``(N, Worst.)'' denotes the average and worst-case accuracy. The best results are highlighted in \textbf{bold}.}
\vspace{-0.13in}
\setlength\tabcolsep{1pt}
\begin{center}
\resizebox{1.0\linewidth}{!}{
\begin{tabular}{l|cccc|cccc|cccc|cccc}
\toprule
    \multirow{3}{*}{\textbf{Method}} & \multicolumn{8}{c|}{\textbf{RGB-Depth Scene Understanding}} & \multicolumn{8}{c}{\textbf{Vision-Language Classification}} \\
    & \multicolumn{4}{c}{\textbf{NYU Depth V2}} & \multicolumn{4}{c|}{\textbf{SUN RGB-D}} & \multicolumn{4}{c}{\textbf{FOOD 101}}  & \multicolumn{4}{c}{\textbf{MVSA}}\\ 
    \cmidrule(lr){2-5} \cmidrule(lr){6-9} \cmidrule(lr){10-13} \cmidrule(lr){14-17}
    & (0,Avg.) & (0,Worst.) & (10,Avg.) & (10,Worst.) 
    & (0,Avg.) & (0,Worst.) & (10,Avg.) & (10,Worst.) 
    & (0,Avg.) & (0,Worst.) & (10,Avg.) & (10,Worst.) 
    & (0,Avg.) & (0,Worst.) & (10,Avg.) & (10,Worst.)  \\ 
\midrule
    CLIP \cite{sun2023eva} & 69.32 & 68.29 & 51.67 & 48.54 & 56.24 & 54.73 & 35.65 & 32.76 & 85.24 & 84.20 & 52.12 & 49.31 & 62.48 & 61.22 & 31.64 & 28.27 \\
    ALIGN \cite{jia2021scaling} & 66.43 & 64.33 & 45.24 & 42.42 & 57.32 & 56.26 & 38.43 & 35.13 & 86.14 & 85.00 & 53.21 & 50.85 & 63.25 & 62.69 & 30.55 & 26.44 \\
    MaPLe \cite{khattak2023maple} & 71.26 & 69.27 & 52.98 & 48.73 & 62.44 & 61.76 & 34.51 & 30.29 & 90.40 & 86.28 & 53.16 & 40.21 & 77.43 & 75.36 & 43.72 & 38.82 \\
    CoOp \cite{jia2022visual} & 67.48 & 66.94 & 49.43 & 45.62 & 58.36 & 56.31 & 39.67 & 35.43 & 88.33 & 85.10 & 55.24 & 51.01 & 74.26 & 73.61 & 42.58 & 37.29 \\
    VPT \cite{jia2022visual} & 62.16 & 61.21 & 41.05 & 37.81 & 54.72 & 53.92 & 33.48 & 29.81 & 83.89 & 82.00 & 51.44 & 49.01 & 65.87 & 64.98 & 32.79 & 29.21 \\
    Late fusion \cite{wang2016learning} & 69.14 & 68.35 & 51.99 & 44.95 & 62.09 & 60.55 & 47.33 & 44.60 & 90.69 & 90.58 & 58.00 & 55.77 & 76.88 & 74.76 & 55.16 & 47.78 \\
    ConcatMML \cite{zhang2021modality} & 70.30 & 69.42 & 53.20 & 47.71 & 61.90 & 61.19 & 45.64 & 42.95 & 89.43 & 88.79 & 56.02 & 54.33 & 75.42 & 75.33 & 53.42 & 50.47 \\
    AlignMML \cite{wang2016learning} & 70.31 & 68.50 & 51.74 & 44.19 & 61.12 & 60.12 & 44.19 & 38.12 & 88.26 & 88.11 & 55.47 & 52.76 & 74.91 & 72.97 & 52.71 & 47.03 \\
    ConcatBow \cite{zhang2023provable} & 49.64 & 48.66 & 31.43 & 29.87 & 41.25 & 40.54 & 26.76 & 24.27 & 70.77 & 70.68 & 35.68 & 34.92 & 64.09 & 62.04 & 45.40 & 40.95  \\
    ConcatBERT \cite{zhang2023provable} & 70.56 & 69.83 & 44.52 & 43.29 & 59.76 & 58.92 & 45.85 & 41.76 & 88.20 & 87.81 & 49.86 & 47.79 & 65.59 & 64.74 & 46.12 & 41.81  \\
    MMTM \cite{joze2020mmtm} & 71.04 & 70.18 & 52.28 & 46.18 & 61.72 & 60.94 & 46.03 & 44.28 & 89.75 & 89.43 & 57.91 & 54.98 & 74.24 & 73.55 & 54.63 & 49.72 \\
    TMC \cite{han2022trusted} & 71.06 & 69.57 & 53.36 & 49.23 & 60.68 & 60.31 & 45.66 & 41.60 & 89.86 & 89.80 & 61.37 & 61.10 & 74.88 & 71.10 & 60.36 & 53.37 \\
    LCKD \cite{wang2023learnable} & 68.01 & 66.15 & 42.31 & 40.56 & 56.43 & 56.32 & 43.21 & 42.43 & 85.32 & 84.26 & 47.43 & 44.22 & 62.44 & 62.27 & 43.52 & 38.63 \\
    UniCODE \cite{xia2023achieving} & 70.12 & 68.74 & 44.78 & 42.79 & 59.21 & 58.55 & 46.32 & 42.21 & 88.39 & 87.21 & 51.28 & 47.95 & 66.97 & 65.94 & 48.34 & 42.95 \\
    SimMMDG \cite{dong2023simmmdg} & 71.34 & 70.29 & 45.67 & 44.83 & 60.54 & 60.31 & 47.86 & 45.79 & 89.57 & 88.43 & 52.55 & 50.31 & 67.08 & 66.35 & 49.52 & 44.01 \\
    MMBT \cite{kiela2019supervised} & 67.00 & 65.84 & 49.59 & 47.24 & 56.91 & 56.18 & 43.28 & 39.46 & 91.52 & 91.38 & 56.75 & 56.21 & 78.50 & 78.04 & 55.35 & 52.22  \\
    QMF \cite{zhang2023provable} & 70.09 & 68.81 & 55.60 & 51.07 & 62.09 & 61.30 & 48.58 & 47.50 & 92.92 & 92.72 & 62.21 & 61.76 & 78.07 & 76.30 & 61.28 & 57.61  \\
    MNL \cite{gong2025multimodal} & 71.52 & 68.75 & 56.72 & 52.15 & 62.25 & 60.45 & 49.10 & 47.80 & 93.03 & 91.45 & 63.16 & 60.50 & 78.85 & 76.80 & 61.82 & 57.90  \\
\midrule
    BOGC-MML & \textbf{74.62} & \textbf{72.15} & \textbf{58.31} & \textbf{54.72}  
    & \textbf{65.48} & \textbf{64.21} & \textbf{50.37} & \textbf{47.93} & \textbf{94.20} & \textbf{93.10} & \textbf{64.10} & \textbf{62.85} 
    & \textbf{79.60} & \textbf{78.90} & \textbf{62.90} & \textbf{59.40} \\
    $\Delta$ (v.s. \text{SOTA}) & $\uparrow$ 3.10 & $\uparrow$ 1.86 & $\uparrow$ 1.51 & $\uparrow$ 2.27 & $\uparrow$ 3.04 & $\uparrow$ 2.45 & $\uparrow$ 1.27 & $\uparrow$ 0.13 & $\uparrow$ 1.05 & $\uparrow$ 0.38 & $\uparrow$ 0.94 & $\uparrow$ 1.09 & $\uparrow$ 0.75 & $\uparrow$ 0.86 & $\uparrow$ 1.08 & $\uparrow$ 1.15 \\
\bottomrule
\end{tabular}
}\end{center}
\label{tab:ex_1}
\end{table*}

\begin{table*}[t]
\centering
\caption{Top-1 accuracy(\%) under missing modalities on the BraTS Dataset. $\bullet$ and $\circ$ denote present and missing modalities, respectively.}
\vspace{-0.1in}
\label{tab:ex_2}
\setlength\tabcolsep{1pt}
\begin{center}
\resizebox{0.95\linewidth}{!}{
\begin{tabular}{cccc|ccccc|ccccc|ccccc}
\toprule
\multicolumn{4}{c|}{\textbf{Modalities}} & \multicolumn{5}{c|}{\textbf{Enhancing Tumour}} & \multicolumn{5}{c|}{\textbf{Tumour Core}} & \multicolumn{5}{c}{\textbf{Whole Tumour}} \\ 
\midrule
Fl & T1 & T1c & T2  
& HMIS & RSeg & mmFm & LCKD & BOGC-MML
& HMIS & RSeg & mmFm & LCKD & BOGC-MML
& HMIS & RSeg & mmFm & LCKD & BOGC-MML \\ 
\midrule
$\bullet$ & $\circ$ & $\circ$ & $\circ$ 
    & 11.78 & 25.69 & 39.33 & 45.48 & \textbf{47.97} 
    & 26.06 & 53.57 & 61.21 & 72.01 & \textbf{75.56} 
    & 52.48 & 85.69 & 86.10 & 89.45 & \textbf{92.36} \\
$\circ$ & $\bullet$ & $\circ$ & $\circ$ 
    & 10.16 & 17.29 & 32.53 & 43.22 & \textbf{46.42} 
    & 37.39 & 47.90 & 56.55 & 66.58 & \textbf{69.21} 
    & 57.62 & 70.11 & 67.52 & 76.48 & \textbf{78.96} \\
$\circ$ & $\circ$ & $\bullet$ & $\circ$ 
    & 62.02 & 67.07 & 72.60 & 75.65 & \textbf{77.86} 
    & 65.29 & 76.83 & 75.41 & 83.02 & \textbf{85.94} 
    & 61.53 & 73.31 & 72.22 & 77.23 & \textbf{79.11} \\
$\circ$ & $\circ$ & $\circ$ & $\bullet$ 
    & 25.63 & 28.97 & 43.05 & 47.19 & \textbf{49.77} 
    & 57.20 & 57.49 & 64.20 & 70.17 & \textbf{73.01} 
    & 80.96 & 82.24 & 81.15 & 84.37 & \textbf{86.95} \\
$\bullet$ & $\bullet$ & $\circ$ & $\circ$ 
    & 10.71 & 32.13 & 42.96 & 48.30 & \textbf{51.21} 
    & 41.12 & 60.68 & 65.91 & 74.58 & \textbf{77.32} 
    & 64.62 & 88.24 & 87.06 & 89.97 & \textbf{93.03} \\
$\bullet$ & $\circ$ & $\bullet$ & $\circ$ 
    & 66.10 & 70.30 & 75.07 & 78.75 & \textbf{81.56} 
    & 71.49 & 80.62 & 77.88 & 85.67 & \textbf{87.42} 
    & 68.99 & 88.51 & 87.30 & 90.47 & \textbf{92.71} \\
$\bullet$ & $\circ$ & $\circ$ & $\bullet$ 
    & 30.22 & 33.84 & 47.52 & 49.01 & \textbf{53.11} 
    & 57.68 & 61.16 & 69.75 & 75.41 & \textbf{78.25} 
    & 82.95 & 88.28 & 87.59 & 90.39 & \textbf{93.78} \\
$\circ$ & $\bullet$ & $\bullet$ & $\circ$ 
    & 66.22 & 69.06 & 74.04 & 76.09 & \textbf{78.98} 
    & 72.46 & 78.72 & 78.59 & 82.49 & \textbf{86.85} 
    & 68.47 & 77.18 & 74.42 & 80.10 & \textbf{83.54} \\
$\circ$ & $\bullet$ & $\circ$ & $\bullet$ 
    & 32.39 & 32.01 & 44.99 & 50.09 & \textbf{52.87} 
    & 60.92 & 62.19 & 69.42 & 72.75 & \textbf{76.02} 
    & 82.41 & 84.78 & 82.20 & 86.05 & \textbf{88.77} \\
$\circ$ & $\circ$ & $\bullet$ & $\bullet$ 
    & 67.83 & 69.71 & 74.51 & 76.01 & \textbf{79.31} 
    & 76.64 & 80.20 & 78.61 & 84.85 & \textbf{86.89} 
    & 82.48 & 85.19 & 82.99 & 86.49 & \textbf{89.72} \\
$\bullet$ & $\bullet$ & $\bullet$ & $\circ$ 
    & 68.54 & 70.78 & 75.47 & 77.78 & \textbf{80.41} 
    & 76.01 & 81.06 & 79.80 & 85.24 & \textbf{88.10} 
    & 72.31 & 88.73 & 87.33 & 90.50 & \textbf{93.24} \\
$\bullet$ & $\bullet$ & $\circ$ & $\bullet$ 
    & 31.07 & 36.41 & 47.70 & 49.96 & \textbf{51.92} 
    & 60.32 & 64.38 & 71.52 & 76.68 & \textbf{79.70} 
    & 83.43 & 88.81 & 87.75 & 90.46 & \textbf{92.82} \\
$\bullet$ & $\circ$ & $\bullet$ & $\bullet$ 
    & 68.72 & 70.88 & 75.67 & 77.48 & \textbf{80.21} 
    & 77.53 & 80.72 & 79.55 & 85.56 & \textbf{87.76} 
    & 83.85 & 89.27 & 88.14 & 90.90 & \textbf{93.37} \\
$\circ$ & $\bullet$ & $\bullet$ & $\bullet$ 
    & 69.92 & 70.10 & 74.75 & 77.60 & \textbf{79.82} 
    & 78.96 & 80.33 & 80.39 & 84.02 & \textbf{87.33} 
    & 83.94 & 86.01 & 82.71 & 86.73 & \textbf{89.11} \\
$\bullet$ & $\bullet$ & $\bullet$ & $\bullet$ 
    & 70.24 & 71.13 & 77.61 & 79.33 & \textbf{83.01} 
    & 79.48 & 80.86 & 85.78 & 85.31 & \textbf{88.61} 
    & 84.74 & 89.45 & 89.64 & 90.84 & \textbf{93.76} \\ 
\bottomrule
\end{tabular}
}
\end{center}
\vspace{-0.15in}
\end{table*}

According to \textbf{Theorem \ref{thm:precision_evidence}}, for the $d$-th dimension, we construct a power mapping $\phi (\cdot)$ to calculate the evidence $e_{t,d}^i$ of $\mathbf{g}_{t}^i$. Then, we compute Dirichlet concentration parameter $\alpha _{t,d}^i$ of $\mathbf{g}_{t}^i$. Also, we can obtain the evidence $e_{t,d}^{\rm fusion}$ and Dirichlet concentration parameter $\alpha _{t,d}^{\rm fusion}$ of $\mathbf{g}_{t}^{\rm fusion}$.
\begin{equation}\label{eq4.3.2}
\begin{aligned}
    e_{t,d}^i=\phi (\boldsymbol{\lambda} ^i_{t,d})=(\boldsymbol{\lambda} ^i_{t,d})^s,&\quad \alpha _{t,d}^i=e_{t,d}^i+1,\\ 
    e_{t,d}^{\rm fusion}=\phi (\boldsymbol{\lambda }_{t,d}^{\rm fusion})=(\boldsymbol{\lambda} ^{\rm fusion}_{t,d})^s,&\quad \alpha _{t,d}^{\rm fusion}=e_{t,d}^{\rm fusion}+1.
\end{aligned}
\end{equation}

Then, according to subjective logic \cite{jsang2018subjective} and evidence theory \cite{bao2021evidential,sensoy2018evidential}, we compute the belief masses (i.e. $b_{t,d}^i$ and $b_{t,d}^{\rm fusion}$) and uncertainties (i.e. $u_t^i$ and $u_t^{\rm fusion}$) of $\mathbf{g}_{t}^i$ and $\mathbf{g}_{t}^{\rm fusion}$ respectively. It can be expressed as follows:
\begin{equation}\label{eq4.3.3}
\resizebox{\linewidth}{!}{$
\begin{aligned}
    &\scalebox{0.9}{$S_t^i=\sum_{d=1}^K \alpha _{t,d}^i,\quad b_{t,d}^i=\frac{e_{t,d}^i}{S_t^i},\quad u_t^i=\frac{K}{S_t^i},$}\\
    &\scalebox{0.9}{$S_t^{\rm fusion}=\sum_{d=1}^K \alpha _{t,d}^{\rm fusion},\quad  b_{t,d}^{\rm fusion}=\frac{e_{t,d}^{\rm fusion}}{S_t^{\rm fusion}},\quad  u_t^{\rm fusion}=\frac{K}{S_t^{\rm fusion}},$}
\end{aligned}
$}
\end{equation}
where $S_t^i$ and $S_t^{\rm fusion}$ are known as Dirichlet strength.
After that, we construct sets of mass $\mathbb{M}^i_t=\{\{b_{t,d}^i\}_{d=1}^K,u_t^i\}$ and $\mathbb{M}^{\rm fusion}_t=\{\{b_{t,d}^{\rm fusion}\}_{d=1}^K,u_t^{\rm fusion}\}$ with the Dempster-Shafer rules, and combine them to obtain a joint mass $\mathbb{M}=\{b_{t,d},u_t\}$ according to \textbf{Eq.27}:
\begin{equation}
\scalebox{0.95}{$b_{t,d}=\frac{b_{t,d}^{i}b_{t,d}^{\rm fusion}+b_{t,d}^{i}u_{t}^{\rm fusion}+b_{t,d}^{\rm fusion}u_{t}^{i}}{1-C},\quad 
u_t=\frac{u_{t}^{i}u_{t}^{\rm fusion}}{1-C}$}
\end{equation}
where $C=\sum _{p\neq q}b_{tp}^{i}b_{tq}^{\rm fusion}$ is a measure of the amount of conflict between the two mass sets.
Then, we can obtain the gradient $\mathbf g_{\rm DS}$ after aggregation as follows:
\begin{equation}\label{eq4.3.4}
    \scalebox{0.95}{$\mathbf g_{\rm DS}=\sum_{t=1}^n\sum _{d=1}^K b_{t,d}(b^i_{t,d}\boldsymbol{\mu }_{t,d}^i+b_{t,d}^{\rm fusion}\boldsymbol{\mu }_{t,d}^{\rm fusion}),$}
\end{equation}
where $\boldsymbol{\mu }_{t,d}^i$ and $\boldsymbol{\mu }_{t,d}^{\rm fusion}$ are the values of $\boldsymbol{\mu }_{t}^i$ and $\boldsymbol{\mu }_{t}^{\rm fusion}$ in the $d$-th dimension, respectively.

To this end, we can obtain a more effective and efficient gradient optimization direction considering the gradient uncertainty across modalities. Briefly, our method first captures the full uncertainty of each gradient source by modeling it as a probability distribution, yielding a richer representation of the gradient space. We then convert each gradient distribution's precision into a scalar evidence measure, which in turn allows us to quantify the uncertainty of every gradient. Finally, we fuse these per-modality uncertainties using a reduced Dempster-Shafer rule to calibrate the combined update direction.
Thereby, our method reduces excessive perturbations in sensitive dimensions and improves update efficiency in insensitive dimensions, leading to more accurate and efficient optimization in MML.

\section{Experiment}
\label{sec:experiment}

In this section, we conduct a series of experiments to evaluate BOGC-MML on various benchmark datasets. 
More details and experiments are provided in \textbf{Appendices D-G}.

\subsection{Experimental Settings}
\label{sec:5.1}

\textbf{Datasets.} 
We evaluate our method on seven datasets covering a range of multi-modal tasks:
(i) audio-visual classification on CREMA-D \cite{cao2014crema} and Kinetics Sounds \cite{arandjelovic2017look};
(ii) RGB-Depth scene understanding on NYU Depth V2 \cite{silberman2012indoor} and SUN RGB-D \cite{song2015sun};
(iii) vision-language classification on UPMC FOOD101 \cite{wang2015recipe} and MVSA \cite{niu2016sentiment};
(iv) multi-modal medical segmentation on BraTS \cite{menze2014multimodal}.

\textbf{Implementation Details.}
Unless otherwise specified, all experiments are conducted using a ResNet-18 backbone as the default architecture configuration, with models trained from scratch without any pre-trained weights. We learn the posterior distribution over the parameters of the final layer in each modality-specific encoder. For optimization, we use stochastic gradient descent (SGD) with a momentum of 0.9 and a scaling factor $\gamma = 1.5$. The initial learning rate is set to 0.1 and can be linearly scaled if necessary. All results are averaged over five runs on NVIDIA Tesla V100 GPUs.

\subsection{Results}
\textbf{Performance Comparison.}
Following \cite{weimmpareto}, we conduct experiments on CREMA-D and Kinetics Sounds datasets. We compare BOGC-MML with representative baselines, e.g., G-Blending \cite{wang2020makes}, OGM-GE \cite{peng2022balanced}, Greedy \cite{wu2022characterizing}, PMR \cite{fan2023pmr}, AGM \cite{li2023boosting}, TMC \cite{han2022trusted}, MMPareto \cite{weimmpareto}, D\&R \cite{wei2024diagnosing}, ARL \cite{wei2025improving}, and ECML \cite{yue2025evidential}.
\textbf{Table \ref{tab:imbalance_cnn}} and \textbf{\ref{tab:imbalance_mbt}} indicate the advantages of BOGC-MML.

\textbf{Performance under Few-Shot and Zero-Shot Settings.}
To evaluate the effectiveness of different fusion strategies, we conduct experiments on four benchmark datasets (NYU Depth V2, SUN RGBD, FOOD101, and MVSA) under 0-shot and 10-shot scenarios. We compare a broad set of representative multi-modal baselines, such as LCKD \cite{wang2023learnable}, UniCODE \cite{xia2023achieving}, SimMMDG \cite{dong2023simmmdg}, MMBT \cite{kiela2019supervised}, QMF \cite{zhang2023provable}, and MNL \cite{gong2025multimodal}.
\textbf{Table \ref{tab:ex_1}} further verify the effectiveness of BOGC-MML.

\textbf{Performance on Missing Modality.}
To further evaluate the performance of BOGC-MML, we conduct comparative experiments under missing modality. We assess BOGC-MML against several strong baselines \cite{bakas2018identifying, wang2023learnable, wang2025causal} across all 15 possible missing modality combinations on the BraTS dataset.
\textbf{Table \ref{tab:ex_2}} demonstrate the advantage of BOGC-MML in practice.

\begin{figure}[t]
    \centering
    \begin{minipage}[t]{0.23\textwidth}
        \includegraphics[width=\textwidth]{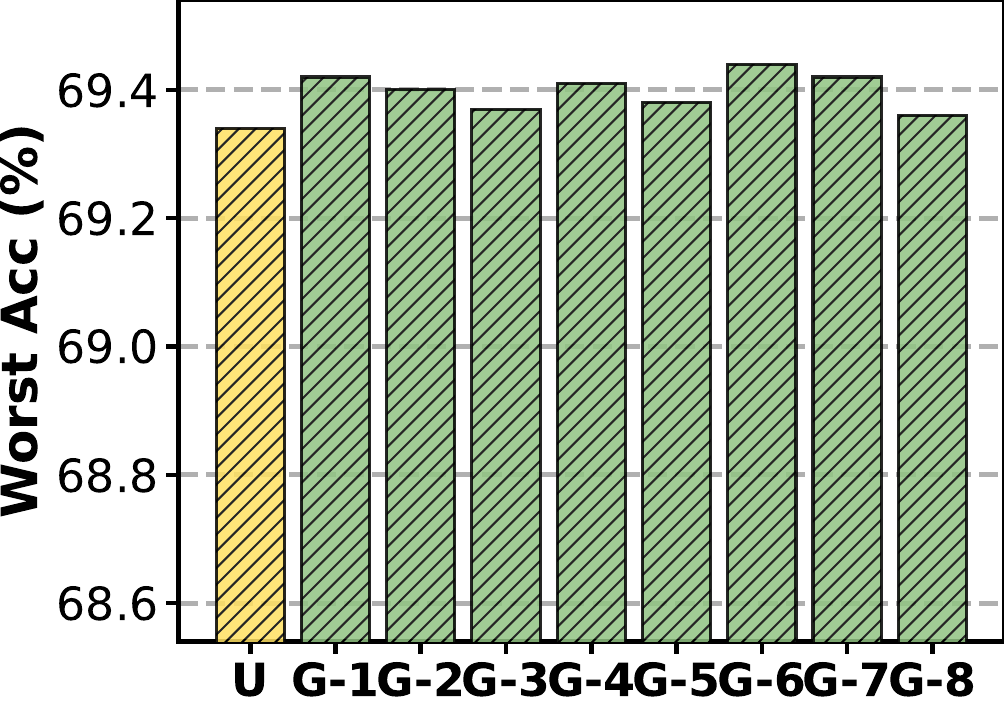}
        \caption{Results with uncertainty in non-conflict scenarios.}
        \label{fig:motivation_uncertainty}
    \end{minipage}\hfill
    \begin{minipage}[t]{0.23\textwidth}
        \includegraphics[width=\textwidth]{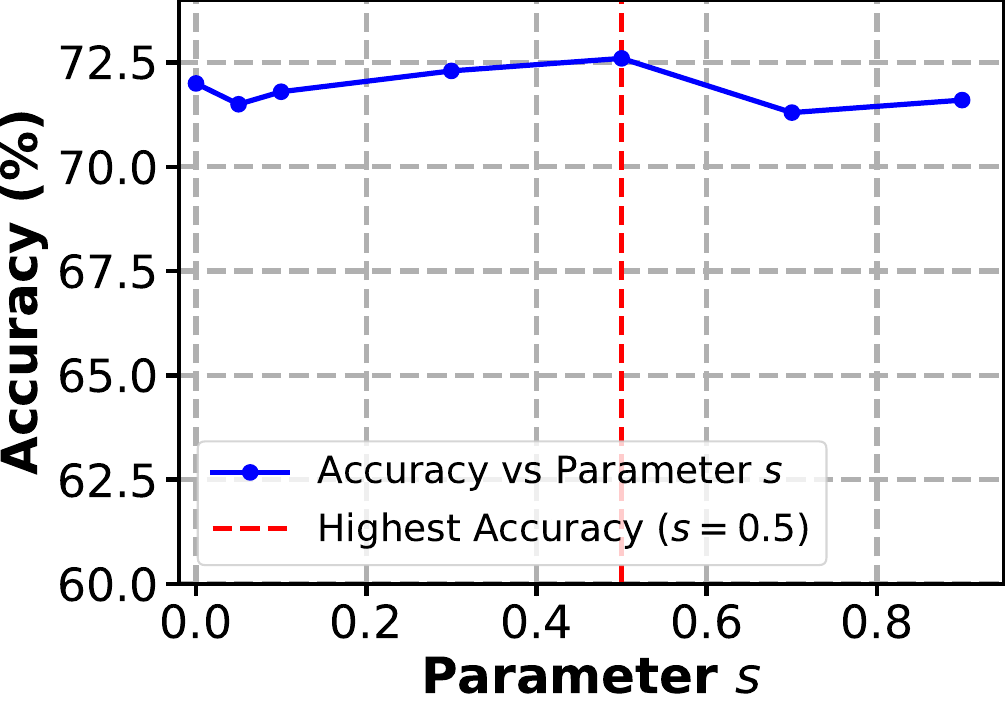}
        \caption{Parameter sensitivity about the parameter $s$.}
        \label{fig:abla_s}
    \end{minipage}
    \vspace{-0.2in}
\end{figure}

\textbf{Ablation Study about}
\textbf{Parameter Sensitivity of Hyperparameter $s$.}
We conduct an experiment to evaluate the impact of $s$ on Kinetics-Sounds, with the range $[0.01, 0.05, 0.1, 0.3, 0.5,0.7, 0.9]$. \textbf{Figure \ref{fig:abla_s}} indicate that (i) when $s=0.5$ achieves the best performance, also our choice in this work; and (ii) the parameter $s$ has a relatively minor influence on the performance.

\section{Conclusion}
\label{sec:conclusion}

This paper rethinks MML from an optimization perspective. Our empirical analysis reveals that performance instability exists even in the absence of gradient conflicts. To account for this, we posit that gradient uncertainty is a critical yet overlooked factor, providing a new explanation for the interesting phenomenon. Building on this argument, we develop BOGC-MML, a framework that shifts the paradigm to explicitly modeling gradient uncertainty. Specifically, we treat gradients as random variables represented by probability distributions. We further design a mapping from gradient precision to evidence, enabling the use of a reduced Dempster's rule to prioritize reliable information during aggregation. 
Extensive experiments on various benchmarks demonstrate the effectiveness and advantages of our approach.

\section*{Impact Statement}
This paper presents work whose goal is to advance the field of 
Machine Learning and Multi-Modal Learning. There are many potential societal consequences of our work, none of which we feel must be specifically highlighted here.

\nocite{langley00}

\bibliography{main}
\bibliographystyle{icml2026}

\end{document}